\documentclass[11pt,a4paper]{article}
\usepackage[hyperref]{naaclhlt2019}
\usepackage{times}
\usepackage{latexsym}
\usepackage{amsthm}
\usepackage{url}
\usepackage{graphicx}
\usepackage{subcaption}
\usepackage{color}
\usepackage{booktabs}

\aclfinalcopy

\usepackage{array}
\newcolumntype{H}{>{\setbox0=\hbox\bgroup}c<{\egroup}@{}}
\title{Matching Entities Across Different Knowledge Graphs\\ with Graph Embeddings}

\author{Michael Azmy, Peng Shi, Jimmy Lin, \and Ihab F. Ilyas\\[0.5ex]
David R. Cheriton School of Computer Science\\
University of Waterloo\\
Waterloo, Ontario, Canada\\[0.5ex]
{\tt \{mwazmy,p8shi,jimmylin,ilyas\}@uwaterloo.ca} \\}

\date{}

\begin{document}
\maketitle
\begin{abstract}
This paper explores the problem of matching entities across different knowledge graphs.
Given a query entity in one knowledge graph, we wish to find the corresponding real-world entity in another knowledge graph.
We formalize this problem and present two large-scale datasets for this task based on exiting cross-ontology links between DBpedia and Wikidata, focused on several hundred thousand ambiguous entities.
Using a classification-based approach, we find that a simple multi-layered perceptron based on representations derived from \textsc{RDF2Vec} graph embeddings of entities in each knowledge graph is sufficient to achieve high accuracy, with only small amounts of training data.
The contributions of our work are datasets for examining this problem and strong baselines on which future work can be based.
\end{abstract}

\section{Introduction}

Knowledge graphs have proven useful for many applications, including document retrieval~\cite{Dalton_etal_SIGIR2014} and question answering~\cite{Mohammed_etal_NAACL2018}.
As there already exist many large-scale efforts such as Freebase~\cite{bollacker2008freebase}, DBpedia~\cite{auer2007dbpedia}, Wikidata~\cite{vrandevcic2014wikidata}, and YAGO~\cite{suchanek2007yago}, to support interoperability there is a need to match entities across multiple resources that refer to the same real-world entity.
Addressing this challenge would, for example, allow mentions in free text that have been linked to entities in one knowledge graph to benefit from knowledge encoded elsewhere.

Ambiguity, of course, is the biggest challenge to this problem:\ for example, there are $21$ persons named ``Adam Smith'' in DBpedia and $24$ in Wikidata.
The obvious solution is to exploit the context of entities for matching.
In this paper, we present two datasets for entity matching between DBpedia and Wikidata that specifically focus on ambiguous cases.
Interestingly, experimental results show that with a classification-based formulation, an off-the-shelf graph embedding, \textsc{RDF2Vec}~\cite{ristoski2016rdf2vec}, combined with a simple multi-layer perceptron (MLP) achieves high accuracy on these datasets.

We view this short paper as having the following two contributions:\ First, we offer the community two large-scale datasets for entity matching, focused on ambiguous entities.
Second, we show that a simple model performs well on these datasets.
Results suggest that \textsc{RDF2Vec} can capture the context of entities in a low dimensional semantic space, and that it is possible to learn associations between distinct semantic spaces (one from each knowledge graph) using a simple MLP to perform entity matching with high accuracy.
Experiments show that only small amounts of training data are required, but a linear model (logistic regression) on the same graph embeddings performs poorly.
Naturally, we are not the first to have worked on aligning knowledge graphs (see discussion in Section~\ref{section:related}).
While more explorations are certainly needed, to our knowledge we are the first to make these interesting observations.

\section{Problem Formulation}
\label{sec:problem}

We begin by formalizing our entity matching problem.
Given a source knowledge graph $S$ containing entities $E^s = \{e_1^s, e_2^s, \dots, e_m^s\}$, where $e_i$ is a Uniform Resource Identifier (URI), for each entity we wish find the entity in the target knowledge graph $T$ containing entities $E^t = \{e_1^t, e_2^t, \dots, e_n^t\}$ that corresponds to the same real-world entity---the common-sense notion that these entities refer to the same person, location, etc.
In our current formulation, we take a query-based approach:\ that is, for a given ``query'' entity in the source knowledge graph, our task is to determine the best matching entity in the target knowledge graph.
Although, in principle, a query entity from the source knowledge graph may be correctly mapped to more than one entity in the target knowledge graph, such instances are rare and we currently ignore this possibility.

To study the entity matching problem, we began by creating two benchmark datasets exploiting \texttt{OWL:sameAs} predicates that link entities between DBpedia (2016-10)\footnote{\url{https://wiki.dbpedia.org/downloads-2016-10}} and Wikidata (2018-10-29).\footnote{\url{https://dumps.wikimedia.org/wikidatawiki/entities/20181029/}}
These predicates are manually curated and can be viewed as high-quality ground truth.
The total number of mappings obtained by querying DBpedia and Wikidata using SPARQL was 6,974,651.
We then removed all mappings referring to Wikipedia disambiguation pages.

Although entities with different names in two knowledge graphs may refer to the same real-world entity, we focus on the ambiguity problem and hence restrict our consideration to entities in knowledge graphs that share the same name---more precisely, the \texttt{foaf:name} predicate in DBpedia and the \texttt{rdfs:label} predicate in Wikidata.
Furthermore, to make our task more challenging, we only consider ambiguous cases (since string matching is sufficient otherwise).
To accomplish this, we first built two inverted indexes of the names of all entities in DBpedia and Wikidata to facilitate rapid querying.
Our problem formulation leads to the construction of two datasets, corresponding to entity matching in each direction:

\smallskip \noindent
\textit{DBpedia to Wikidata:}
Here, we take DBpedia as the source knowledge graph and Wikidata as the target.
For each entity in DBpedia, we queried the above index to retrieve entities with the same name in Wikidata, which forms a candidate set for disambiguation.
Since our focus is ambiguous entities, we discard source DBpedia entities in which there is only one entity with the same name in Wikidata.
For example, there are several people with the name ``John Burt'':\ John Burt (footballer), John Burt (rugby union), John Burt (anti-abortion activist), and John Burt (field hockey).
Of these, only one choice is correct, which is determined by the \texttt{owl:sameAs} predicate:\ this provides our positive ground truth label.
Thus, by construction in each candidate set there is \textit{only} one positive candidate and at least one negative candidate.
This dataset contains 376,065 unique DBpedia URIs comprising the queries with a total of 232,757 unique names, and 967,937 unique Wikidata URIs as candidates.

\smallskip \noindent
\textit{Wikidata to DBpedia:}
We can apply exactly the same procedure as above to build a dataset with Wikidata as the source and DBpedia as the target.
The resulting dataset contains 329,320 unique Wikidata URIs comprising the queries with a total of 293,712 unique names and 523,517 unique DBpedia URIs as candidates.

\begin{figure}
	\centering
	\begin{subfigure}[b]{0.49\linewidth}
		\includegraphics[width=\textwidth]{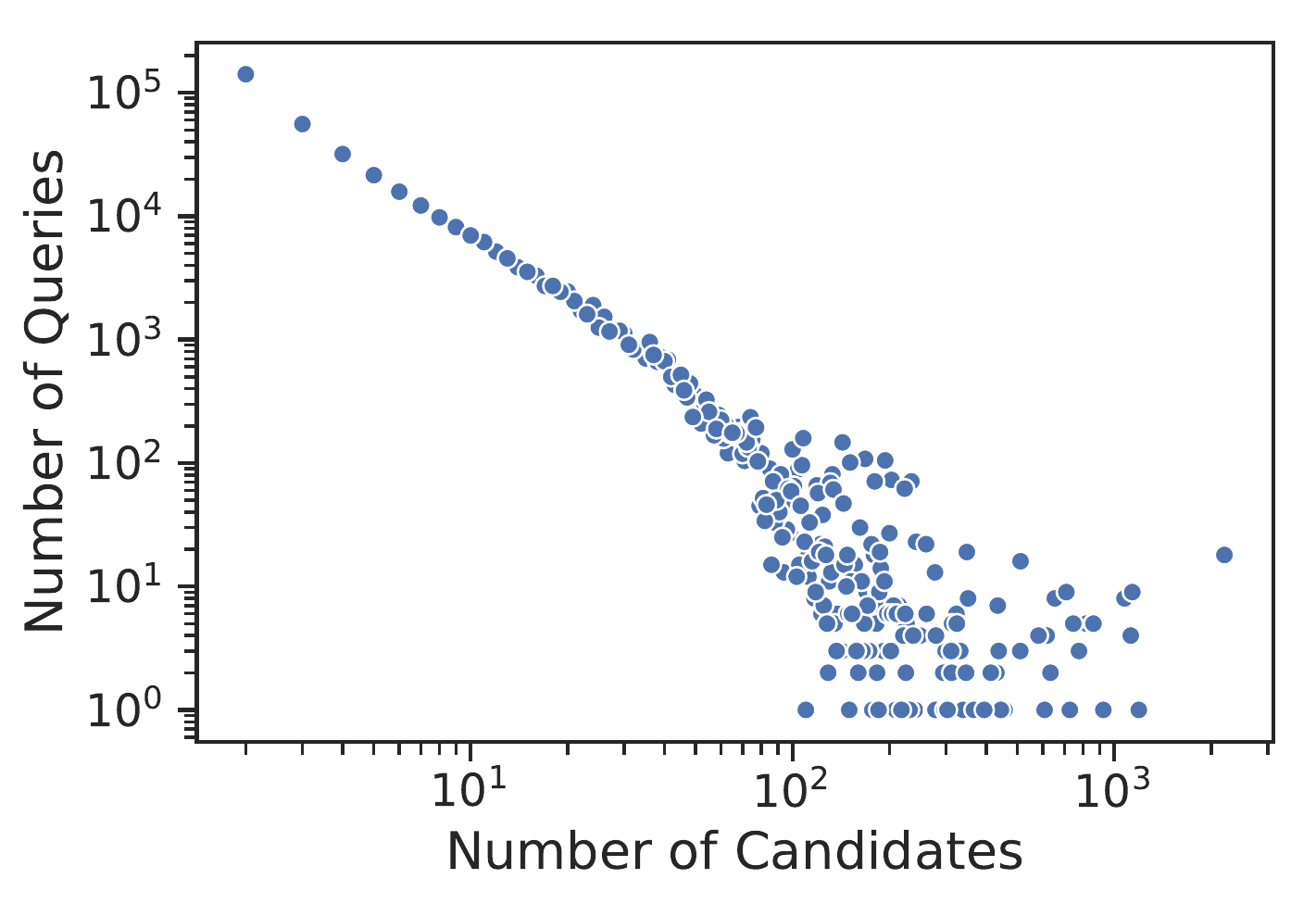}
		\caption{DBpedia to Wikidata}
		\label{fig:scatter_db}
	\end{subfigure}
	\begin{subfigure}[b]{0.49\linewidth}
		\includegraphics[width=\textwidth]{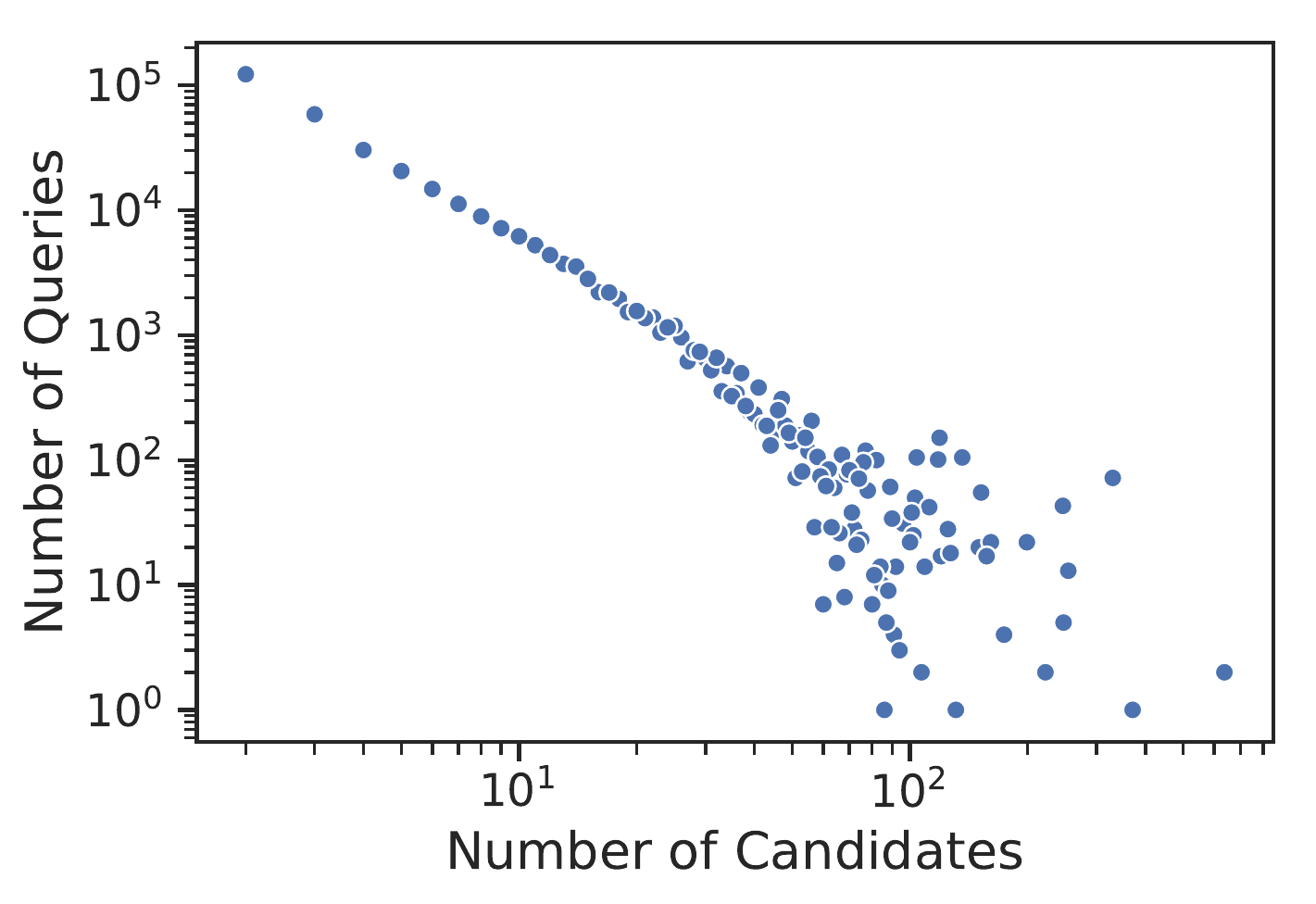}
		\caption{Wikidata to DBpedia}
		\label{fig:scatter_wd}
	\end{subfigure}
	\caption{Log-log plot showing the number of candidate URIs for each query URI.}\label{fig:scatter}
\end{figure}

\medskip \noindent
Finally, we shuffle and split the data into training, validation, and test sets with a ratio of 70\%, 10\%, and 20\%, respectively. 
Statistics are summarized in Table~\ref{tab:stats}, and we make these datasets publicly available.\footnote{\url{https://github.com/MichaelAzmy/ambiguous-dbwd}}
Figure~\ref{fig:scatter} shows the number of query (source) entities with different numbers of candidate (target) entities.
We see Zipf-like distributions:\ although most query entities have only a modest number of candidates, there exist outliers with hundreds or more candidates.

Note that by construction, our datasets for evaluating entity matching {\it cannot} be solved by NLP techniques based on text alone, since the source entities and target entities share exactly the same name (thus string matching conveys no information).
The context for disambiguation must come from some non-text source (in our case, captured in graph embeddings).

\begin{table}[t]
\centering
\begin{tabular}{lcccc}
\toprule
  Dataset &Training & Validation & Testing & \textbf{Total}\\
  \toprule
  DB to WD & 263,245 &  37,607 & 75,213 & \textbf{376,065} \\
  WD to DB & 230,523 & 32,933  & 658,64 & \textbf{329,320} \\
\bottomrule
\end{tabular}
\caption{Statistics of the datasets.}
\label{tab:stats}
\end{table}

\section{Classification Model}

We propose a classification approach to tackle the entity matching problem across knowledge graphs.
Here, we use a point-wise training strategy:\ a classifier is trained on each source--target entity pair and the probability of predicting a match is used for candidate ranking.

A graph embedding is used to represent nodes of a graph in some low dimensional semantic space while preserving some aspect of its structure.
Different graph embedding techniques have been introduced recently to capture different aspects of the graph structure.
In our case, we need to preserve the structural as well as the semantic features of the nodes (entities) so that semantically-similar nodes are close to each other in the embedding space.
For this, we decided to use the \textsc{RDF2Vec}~\cite{ristoski2016rdf2vec} graph embedding technique. 

In \textsc{RDF2Vec}, the RDF graph is first ``unfolded'' into a set of $k$ sequences of entities with predicates connecting them, forming natural language sentences. 
This is typically performed using two approaches:\ graph walks and Weisfeiler-Lehman Subtree RDF graph kernels.
After that, the generated sentences are used to train a Word2Vec~\cite{mikolov2013efficient, mikolov2013distributed} model over the natural language output. 
The outcome of this step is a \textit{d}-dimensional vector for each entity (i.e., node in the knowledge graph).

After the above process, the embedding of the query entity in the source knowledge graph and the candidate entity in the target knowledge graph are concatenated into one feature vector and then fed into a multi-layer perception with one hidden layer using the ReLU activation function, followed by a fully-connected layer and softmax to output the final prediction.
The model is trained using the Adam optimizer~\cite{kingma2014adam}, and negative log-likelihood loss is used.
Each pair of training example is associated with the ground truth from the datasets described in the previous section.
We rank the candidates by the match probability for evaluation.
As a baseline, we compare our MLP with a simple logistic regression (LR) model over the same input vectors.

\section{Experimental Evaluation}
\label{sec:exp}

We use the datasets introduced in Section~\ref{sec:problem} to evaluate our model.
For each query, there is one positive candidate and several negative candidates.
Entities in DBpedia and Wikidata are embedded independently, and thus matching entities have two different embeddings, one in each knowledge graph.
We use pretrained embeddings\footnote{\url{http://data.dws.informatik.uni-mannheim.de/rdf2vec/}} with hyperparameters $k = 200$ walks and depth $l = 4$. 
The embeddings were trained using the skip-gram model with $d = 500$, which showed good results in~\citet{ristoski2016rdf2vec}.
If an entity has no pretrained embedding, a randomly-initialized vector is used. 

We evaluate the model on the test set with the best configuration tuned on the validation set, using the entire training set.
The MLP hidden layer has size 750, with a learning rate of $0.001$ and a batch size of $1024$. Mean reciprocal rank (MRR) is used for evaluation.

\begin{figure}
	\centering
	\begin{subfigure}[b]{0.49\linewidth}
		\includegraphics[width=\textwidth]{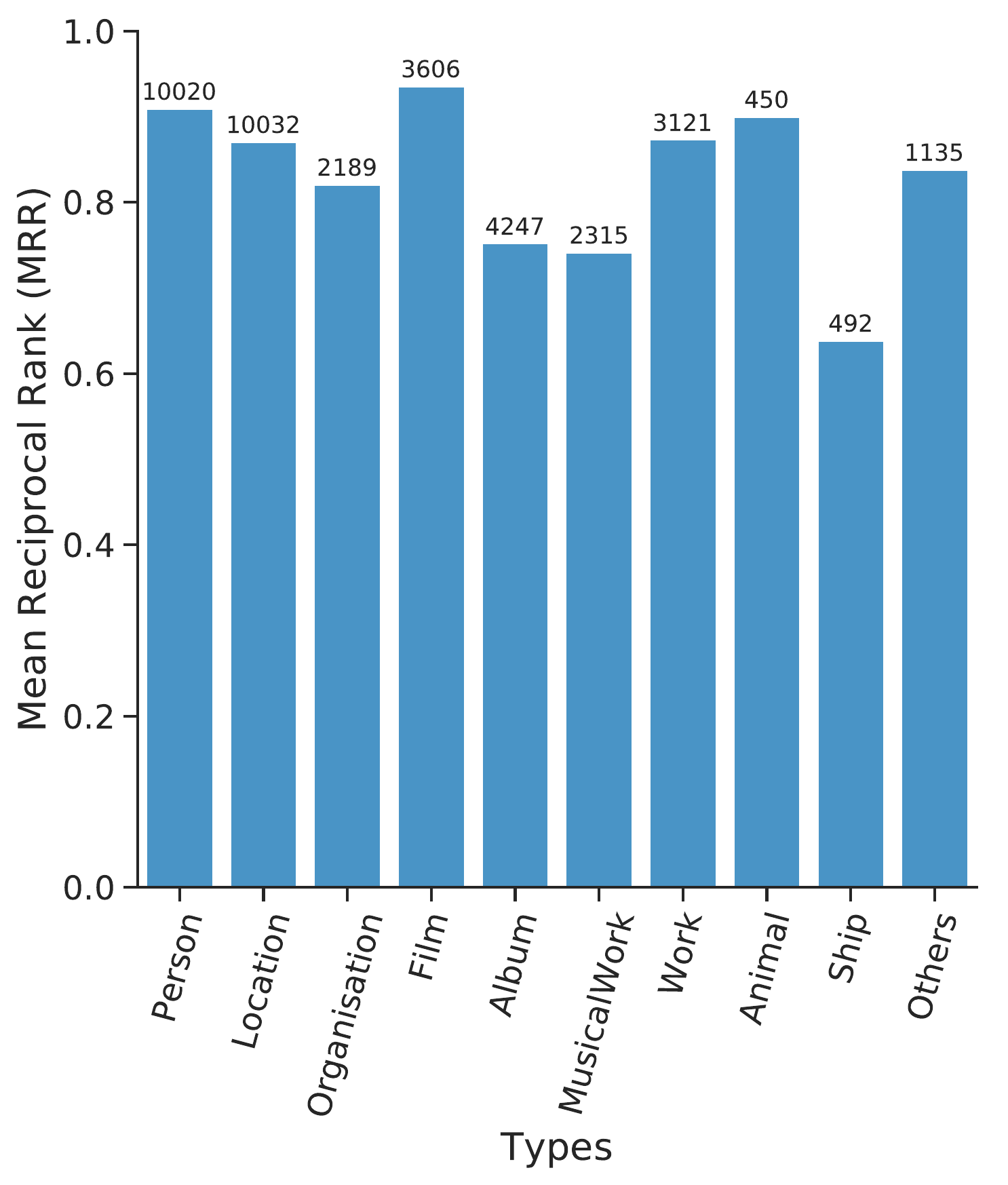}
		\caption{DBpedia to Wikidata}
		\label{fig:db_wd_types}
	\end{subfigure}
	\begin{subfigure}[b]{0.49\linewidth}
		\includegraphics[width=\textwidth]{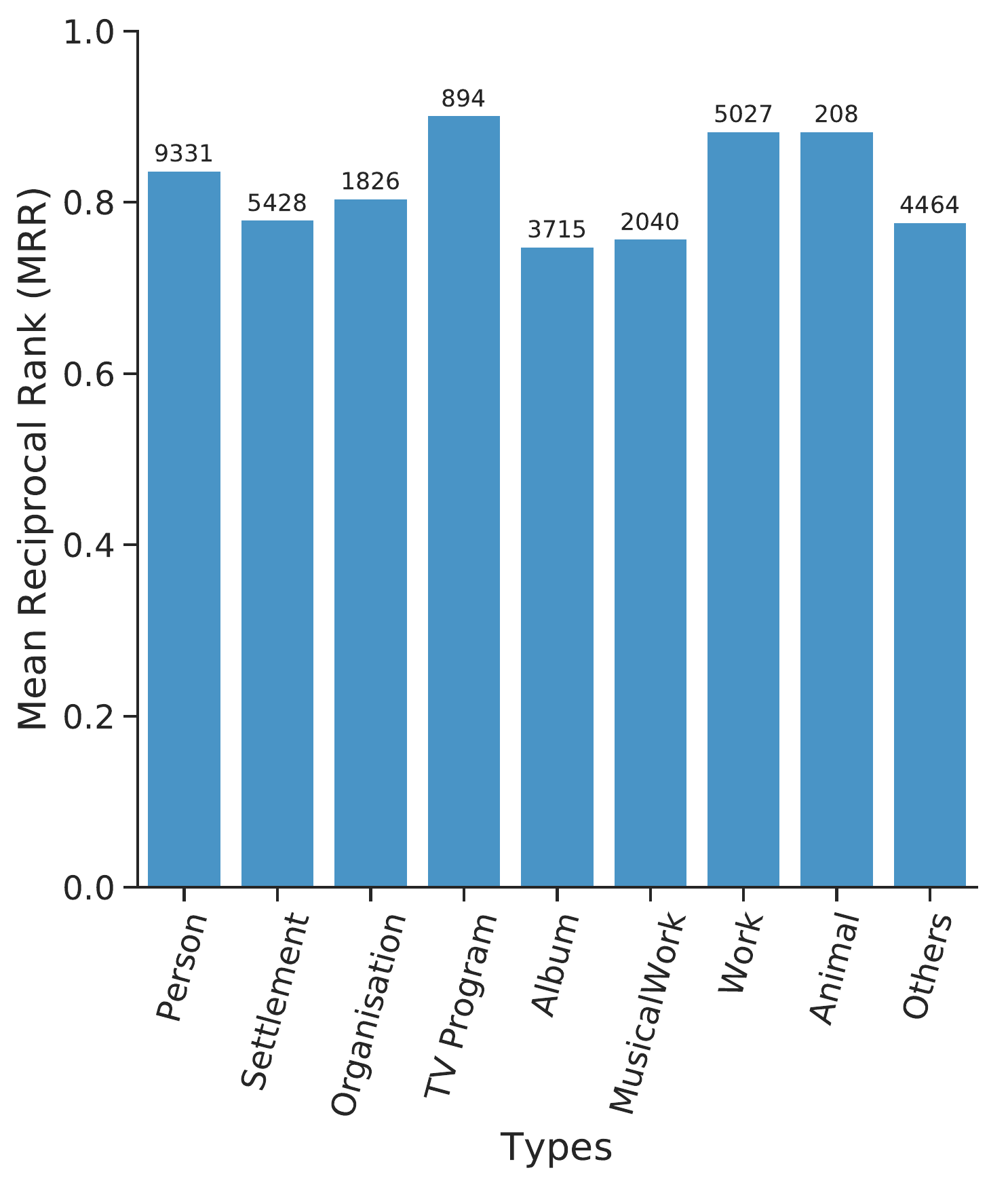}
		\caption{Wikidata to DBpedia}
		\label{fig:wd_db_types}
	\end{subfigure}
	\caption{Mean Reciprocal Rank (MRR) on the validation sets for different types of queries. Number above each bar indicates number of query entities.}
	\label{fig:types}
\end{figure}

Overall, on the test set, the MLP achieves 0.85 MRR matching DBpedia entities to Wikidata and 0.81 MRR matching Wikidata entities to DBpedia. 
In comparison, logistic regression fails to learn a good decision boundary and achieves only 0.64 MRR and 0.62 MRR, respectively. 
Note that the embeddings of each knowledge graph are learned separately, which means that our model is {\it not} simply learning to match words in semantic relations---but actually learning correspondences between two semantic spaces.
For reference, a random guessing baseline yields 0.25 MRR and 0.32 MRR, respectively. 

Figure~\ref{fig:types} breaks down MRR according to entity type.
We observe that types \textit{Album} and \textit{MusicalWork} yield worse results than the others, primarily because of greater ambiguity.
These two types have larger candidates sizes, averaging 12.1 and 11.0 respectively, compared to persons, whose average candidates size is only 6.5 for the DBpedia-to-Wikidata dataset.

\begin{figure}
	\centering
	\begin{subfigure}[b]{0.49\linewidth}
		\includegraphics[width=\textwidth]{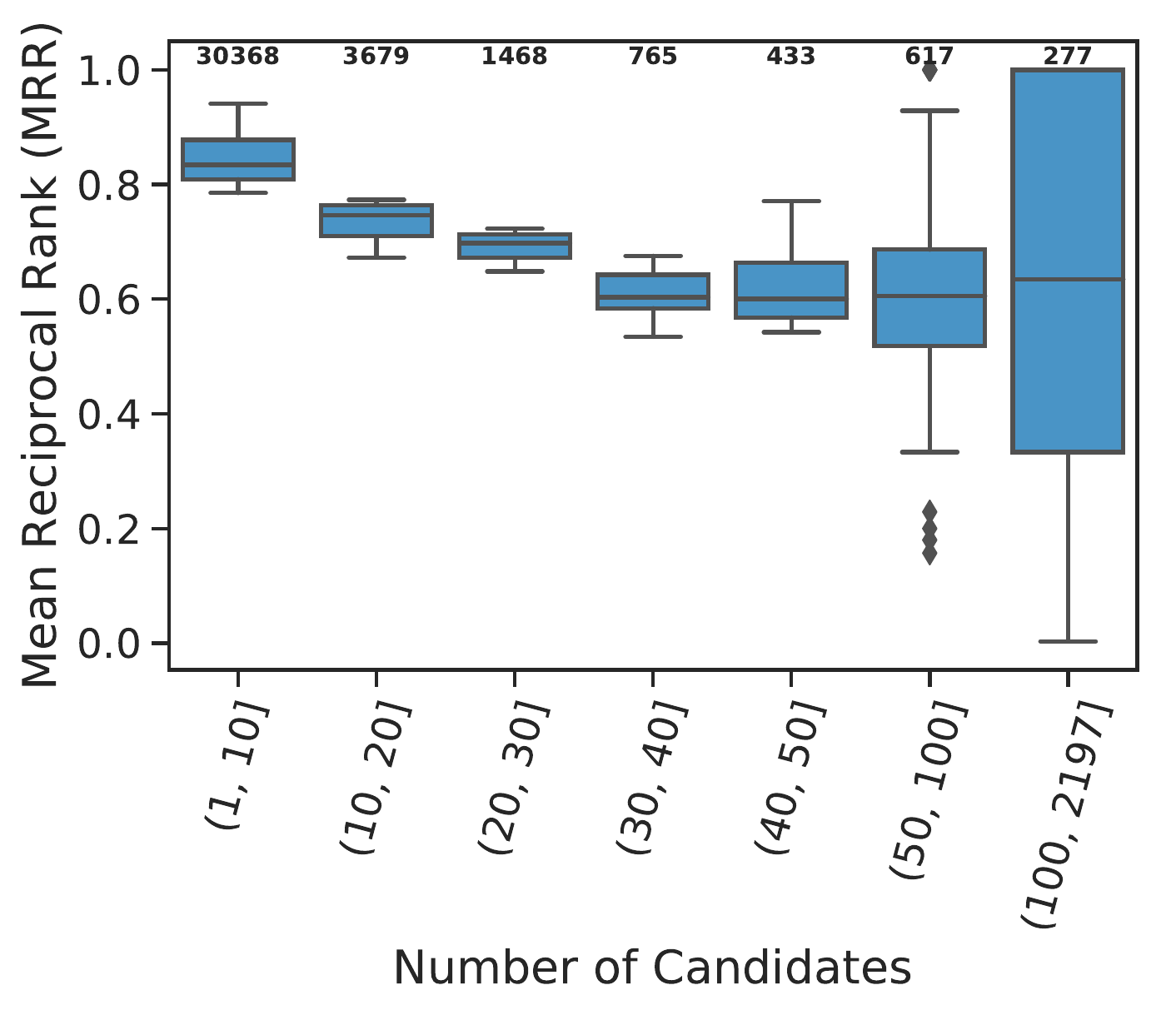}
		\caption{DBpedia to Wikidata}
		\label{fig:db_wd_mrr}
	\end{subfigure}
	\begin{subfigure}[b]{0.49\linewidth}
		\includegraphics[width=\textwidth]{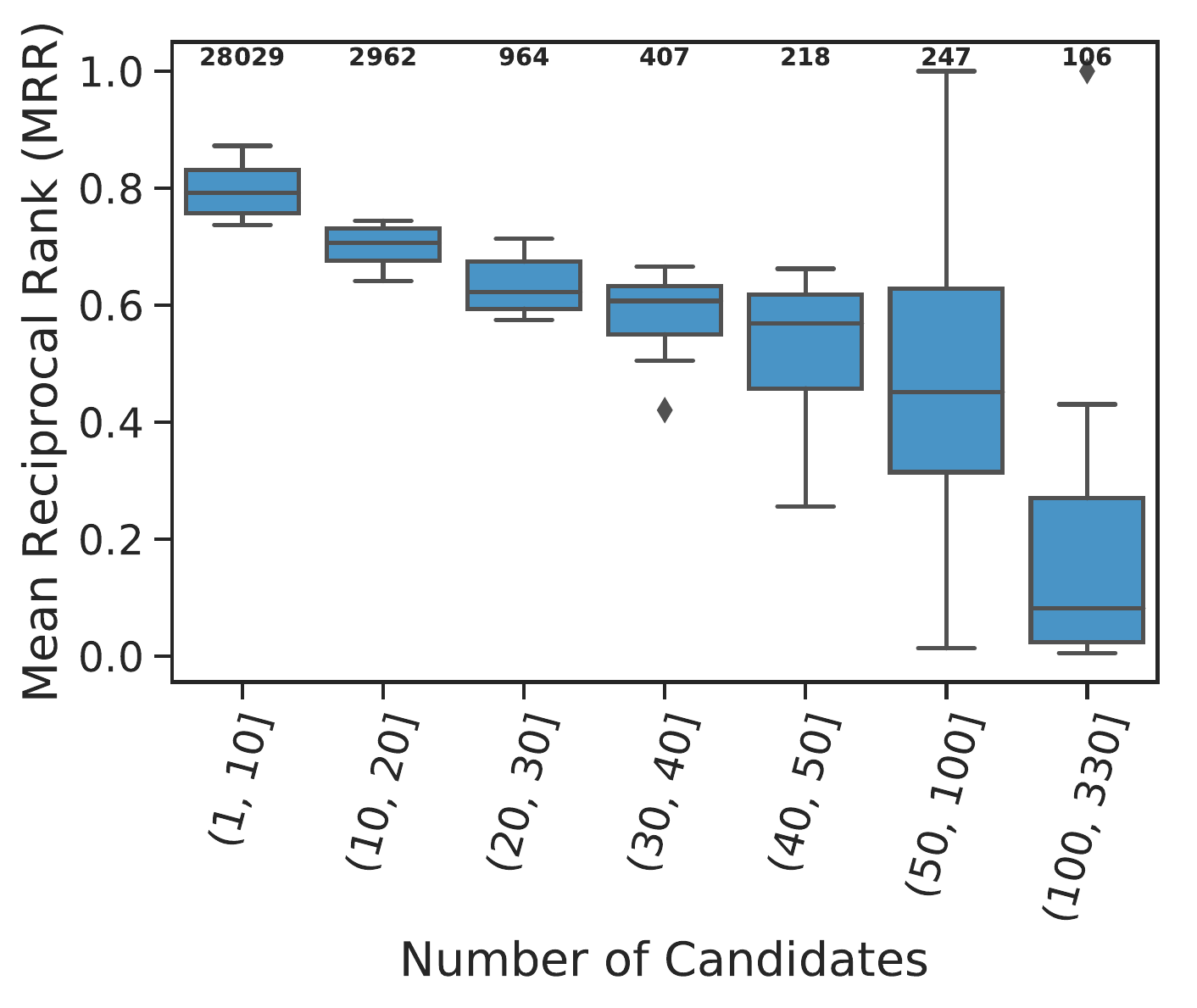}
		\caption{Wikidata to DBpedia}
		\label{fig:wd_db_mrr}
	\end{subfigure}
	\caption{Mean Reciprocal Rank (MRR) on validation sets for different numbers of candidates. Number above each bar indicates number of queries in that bin.}
	\label{fig:mrr}
\end{figure}

Based on error analysis, we observe that our model lacks the fine-grained ability to disambiguate entities in the same type/domain in some cases.
For example, in the \textit{music} domain, our model cannot differentiate the record company \textit{Sunday Best} and the single \textit{Sunday Best} by Megan Washington.
Overall, in the validation set of the DBpedia-to-Wikidata dataset, for cases where the model fails to place the correct entity at rank one but succeeds at rank two instead, 79.5\% of cases have the same type of entity in both positions.

In our next analysis, we investigate the effect of the size of the candidate sets against matching accuracy.
We measure the MRR of the MLP model on the validation sets, broken down into buckets according to different numbers of candidates, summarized as boxplots.
This is shown in Figure~\ref{fig:mrr}:\ as expected, MRR decreases overall as the number of candidates increases.

\begin{figure}[t]
	\centering
	\begin{subfigure}[b]{0.49\linewidth}
		\includegraphics[width=\textwidth]{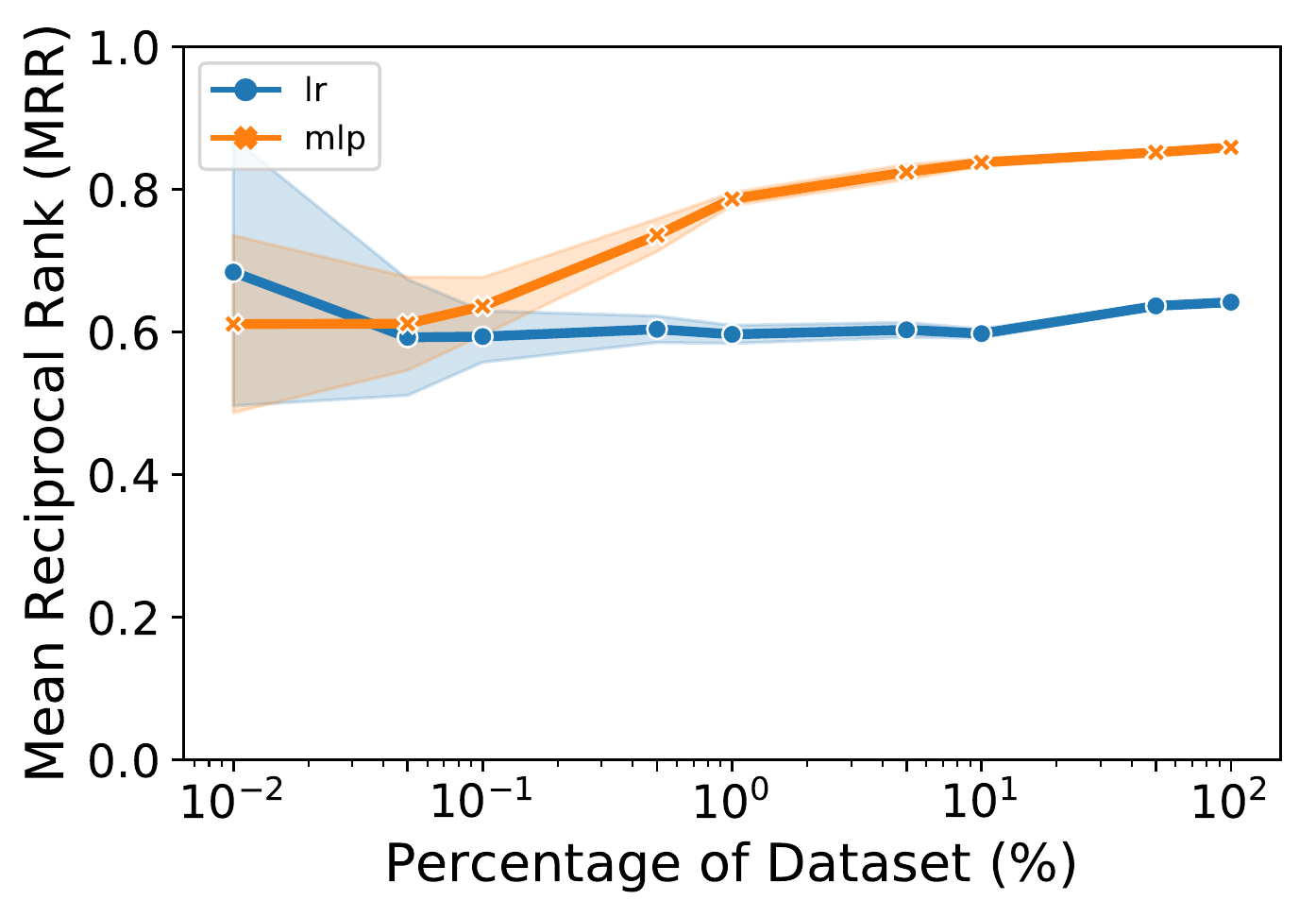}
		\caption{DBpedia to Wikidata}
		\label{fig:db_wd_train}
	\end{subfigure}
	\begin{subfigure}[b]{0.49\linewidth}
		\includegraphics[width=\textwidth]{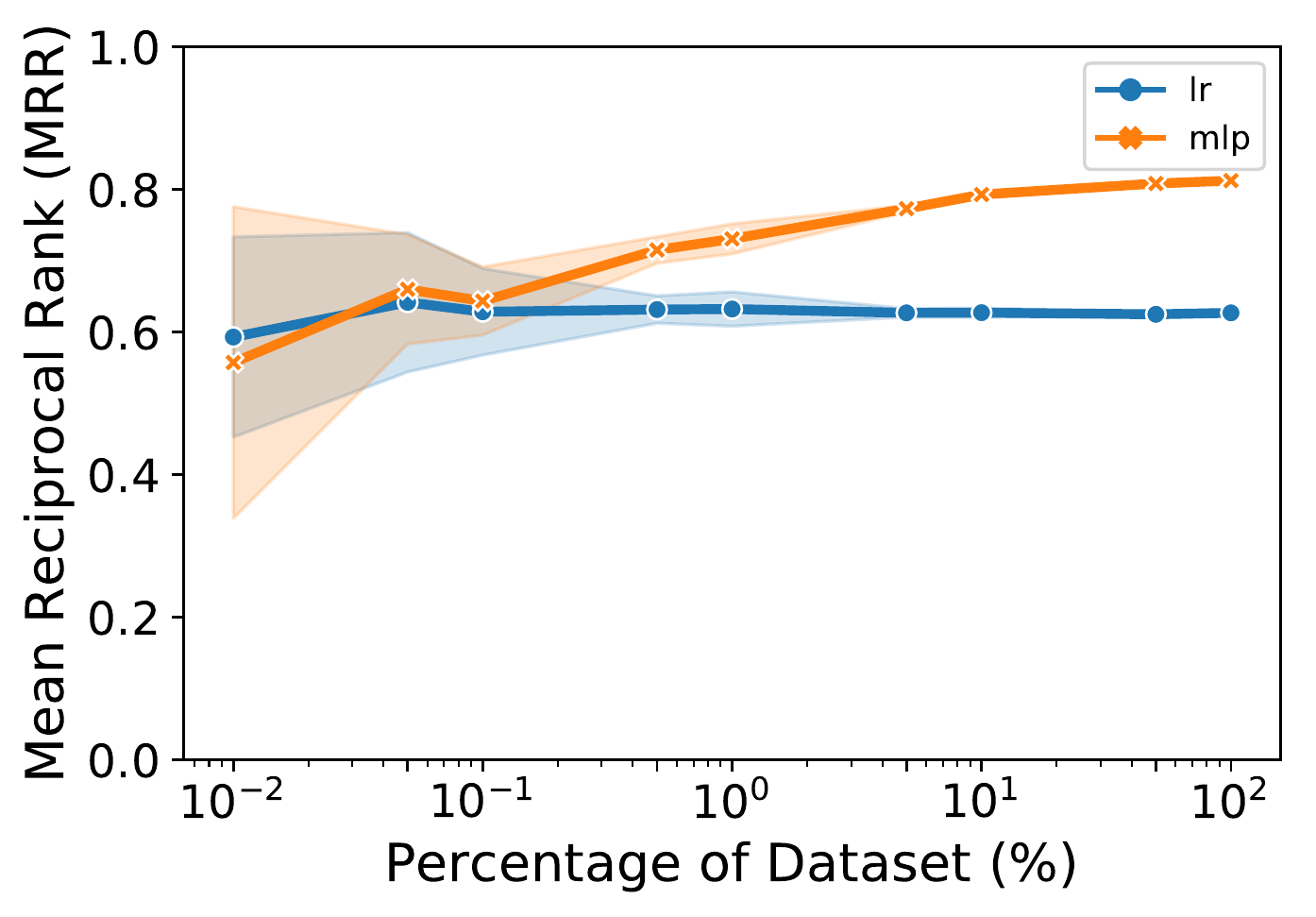}
		\caption{Wikidata to DBpedia}
		\label{fig:wd_db_train}
	\end{subfigure}
	\caption{Mean Reciprocal Rank (MRR) when varying the size of the training set ($x$ axis is in log scale).}
	\label{fig:train}
\end{figure}

Finally, we wish to examine the effects of training data size.
Figure~\ref{fig:train} shows the effects of changing the size of the training set with percentages $\in \{0.01, 0.05, 0.1, 0.5, 1, 5, 10, 50, 100\}$ on the MRR evaluated on the validation sets.
In each case, we randomly divide the sampled data into training/validation splits while preserving the 70:10 ratio.
We repeat the sampling and run the models $10$ times for the first four points and $5$ times for the rest; 95\% confidence intervals are shown in the plot as shaded regions.
We observe that with a small amount of training data, the MLP model can achieve reasonable MRR.
For example, with only $0.5\%$, the MLP model can achieve $0.82$ and $0.77$ MRR, compared to $0.85$ and $0.81$ on the entire training set in the DBpedia-to-Wikidata and Wikidata-to-DBpedia datasets, respectively.

\section{Related Work}
\label{section:related}

Research related to knowledge graph integration comes from the database community and focuses on ontology matching---referred to as record-linkage, entity resolution, or deduplication. 
Examples include \textsc{Magellan}~\cite{konda2016magellan}, \textsc{DeepER}~\cite{ebraheem2018distributed}, and the work of \citet{mudgal2018deep}.
The primary difference between this work and ours is that they assume relational structure and that the tables to be matched have been already aligned using schema matching techniques.
These systems cannot be directly applied to entity matching across knowledge graphs due to differences in structure between the relational model and the RDF model. 

The Semantic Web community has studied the problem of matching entities across knowledge graphs, for example, the Ontology Alignment Evaluation Initiative (OAEI) on ontology matching in knowledge graphs. 
However, the benchmarks used in these evaluations are quite small.
For example, the \textit{spimbench} benchmark~\cite{saveta2015pushing} has a total of only 1800 instances and 50,000 triples.

According to~\citet{castano2011ontology}, entity matching on knowledge graphs can be classified into:\ 
(1) value-oriented approaches that define the similarity between instances on the attribute level and an appropriate matching technique is used based on attribute type, and 
(2) record-oriented approaches which include learning-based, similarity-based, rule-based, and context-based techniques. 

The best approaches in OAEI 2017 either rely on logical reasoning as in~\citet{jimenez2011logmap} or on textual features as in~\citet{achichi2017legato}.
In contrast, our work differs in the following ways:\
(1) we use graph embeddings to capture the semantics and structure of the knowledge graphs without the need for hand-crafted features,
(2) our system does not require any schema mappings, and
(3) our approach can take advantage of the graph nature of RDF, including knowledge about connectivity between nodes and how they relate to one another.

\section{Conclusions}

We explore the problem of entity matching across knowledge graphs, sharing with the community two benchmark datasets and a baseline model.
Although quite simple, our model reveals some insights about the nature of this problem and paves the way for future work.

\section*{Acknowledgments}

This research was supported by the Natural Sciences and Engineering Research Council (NSERC) of Canada.


\end{document}